\documentclass[conference]{IEEEtran}
\IEEEoverridecommandlockouts

\usepackage{cite}
\usepackage{amsmath,amssymb,amsfonts}
\usepackage{algorithmic}
\usepackage{graphicx}
\usepackage{textcomp}
\usepackage{xcolor}
\usepackage{booktabs} 
\usepackage{array}
\usepackage{tabularx}
 
\usepackage{caption}

\renewcommand{\thetable}{\Roman{table}}  

\def\BibTeX{{\rm B\kern-.05em{\sc i\kern-.025em b}\kern-.08em
    T\kern-.1667em\lower.7ex\hbox{E}\kern-.125emX}}
\begin{document}

\title{HyDRA: Hierarchical and Dynamic Rank
Adaptation for Mobile Vision Language Model\\

}
\author{
    \IEEEauthorblockN{Yuanhao Xi$^{1,2,3}$\IEEEauthorrefmark{2}, Xiaohuan Bing$^{1,2,3}$\IEEEauthorrefmark{2}, Ramin Yahyapour$^{2,3}$\IEEEauthorrefmark{1}}
    \IEEEauthorblockA{$^1$ Liaoning Technical University, Huludao, China}
    \IEEEauthorblockA{$^2$ University of Göttingen, Göttingen, Germany}
    \IEEEauthorblockA{$^3$ Gesellschaft für Wissenschaftliche Datenverarbeitung mbH Göttingen, Göttingen, Germany}
      \thanks{\IEEEauthorrefmark{2}Equal contribution }
    \thanks{ Emails: \{yuanhao.xi, xiaohuan.bing\}@stud.uni-goettingen.de }\thanks{\IEEEauthorrefmark{1}Corresponding author: Ramin Yahyapour(ramin.yahyapour@gwdg.de)}

}


\maketitle

\begin{abstract}
Vision Language Models (VLMs) have undergone significant advancements, particularly with the emergence of mobile-oriented VLMs, which offer a wide range of application scenarios. However, the substantial computational requirements for training these models present a significant obstacle to their practical application. To address this issue, Low-Rank Adaptation (LoRA) has been proposed. Nevertheless, the standard LoRA with a fixed rank lacks sufficient capability for training mobile VLMs that process both text and image modalities. In this work, we introduce HyDRA, a parameter-efficient fine-tuning framework designed to implement hierarchical and dynamic rank scheduling for mobile VLMs. This framework incorporates two essential optimization strategies: (1) hierarchical optimization, which involves a coarse-grained approach that assigns different ranks to various layers, as well as a fine-grained method that adjusts ranks within individual layers, and (2) dynamic adjustment, which employs an end-to-end automatic optimization using a lightweight performance model to determine and adjust ranks during the fine-tuning process. Comprehensive experiments conducted on popular benchmarks demonstrate that HyDRA consistently outperforms the baseline, achieving a 4.7\% improvement across various model sizes without increasing the number of trainable parameters. In some tasks, it even surpasses full-parameter fine-tuning.
\end{abstract}


\section{Introduction}

In recent years, the field of multimodal large language models experiences rapid development, providing novel solutions to address complex tasks spanning various modalities~\cite{yin2023survey,li2024vision}. Some researchers have already applied VLMs to mobile devices, such as MobileVLM~\cite{chu2023mobilevlm}. This significantly broadens the application scenarios of VLMs and possesses considerable practical value. However, training VLMs requires substantial computational resources~\cite{luo2024cheap,zhang2024vision}. Therefore, developing an efficient fine-tuning methodology specifically tailored for mobile-oriented VLMs is of significant practical importance.

 To bridge this gap, techniques such as LoRA~\cite{hu2021lora} have been developed. LoRA and its variants have demonstrated excellent performance in the fine-tuning tasks of current large language models (LLMs). The variants of LoRA can be divided into three categories: the first involves dynamically adjusting the rank values of LoRA, the second focuses on enhancements to LoRA, and the third involves using multiple LoRA~\cite{han2024parameter,xu2023parameter}. The DyLoRA~\cite{valipour2022dylora} sorts the representations learned by the adapter module at various ranks during fine-tuning, enabling the LoRA blocks to be trained across a range of ranks rather than a fixed rank. QLoRA~\cite{dettmers2024qlora} introduces a 4-bit NormalFloat data type to save memory without sacrificing performance. LoRAHub~\cite{huang2023lorahub} gathers diverse task-specific LoRA modules and autonomously combines suitable ones without human input. However, these methods, which are applied to LLMs, are not suitable for the instruction tuning phase of mobile VLMs. Unlike LLMs, which primarily handle text, mobile VLMs process a combination of text and image modalities. Due to the differing sensitivities of text and image modalities to various layers of   the model during fine-tuning, the final performance on downstream tasks is unsatisfactory.

To address these issues, we propose adopting different rank settings for different layers to account for the varying sensitivities of multimodal tasks. As shown in Fig. 1,we introduce HyDRA, which incorporates two key techniques: hierarchical optimization and dynamic adjustment for the rank.
\vspace{-1em} 
 \begin{figure}[t]
\centering
\includegraphics[width=0.95\linewidth]{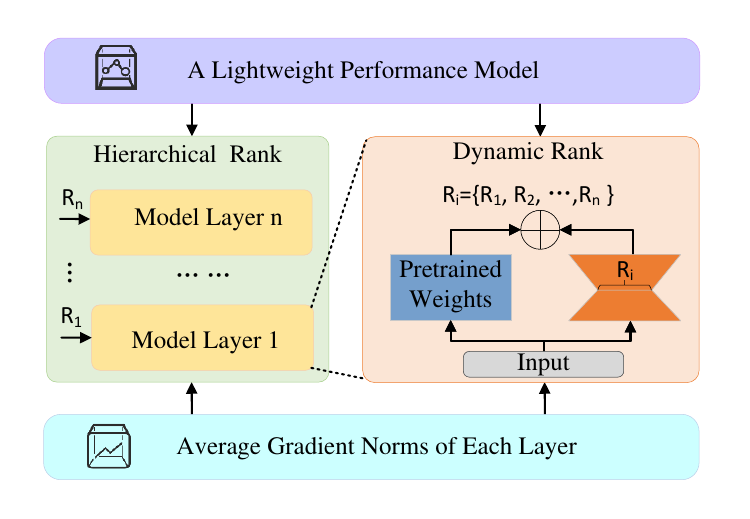}
\caption{An illustration of hierarchical and dynamic rank adaptation.  The average gradient norms serve as the basis for assigning the rank of layers. A lightweight performance model determines the optimal set of rank values.}
\label{fig:1}
\end{figure}
\begin{figure*}[ht]
\centering
\includegraphics[width=\linewidth]{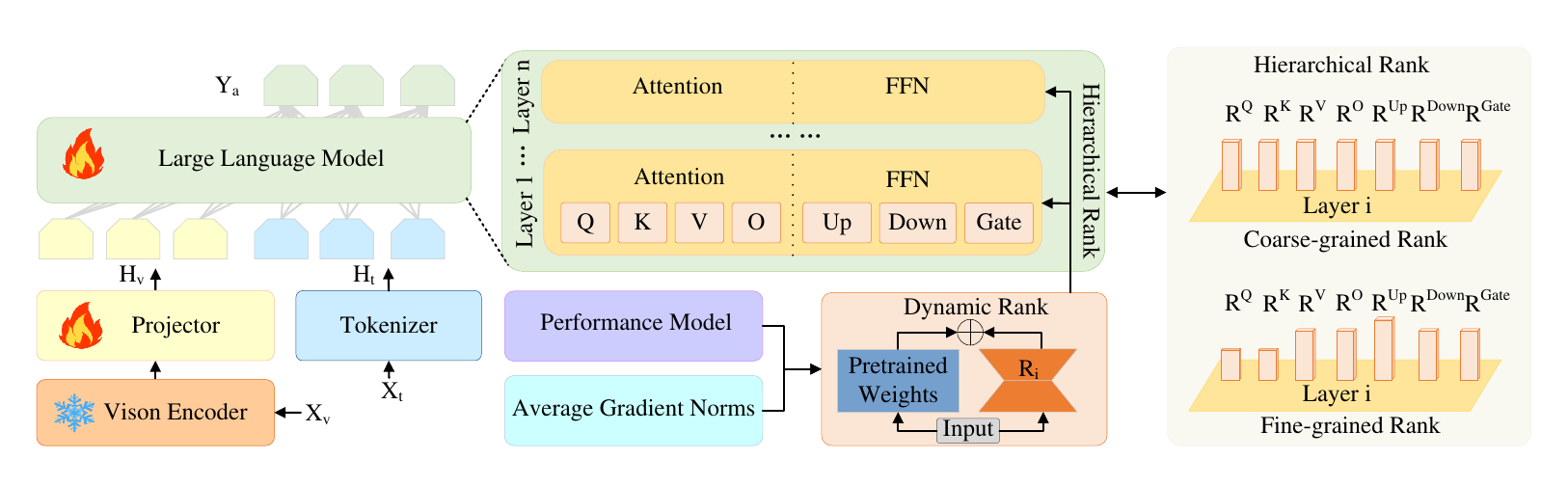}
\caption{Overview of HyDRA with hierarchical rank optimization and dynamic adjustment for mobile VLMs instruction tuning. $X_t$ and $X_v$ represent the text and image tokens, respectively. $R^{Up}$ represents the rank for the up projection of the feed-
forward neural network (FFN). $R^Q$, $R^K$, $R^{Gate}$, etc., are similarly defined. }
\end{figure*}

In hierarchical optimization, we refine the rank values of different layers based on their uneven distribution of average gradient norms. In dynamic adjustment, we utilize a lightweight performance model to allocate different ranks to each layer, thereby enhancing performance on downstream tasks. The overview of our framework can be illustrated in Fig. 2.

We conduct comprehensive experiments, and the results demonstrate that HyDRA performs best in most scenarios. The contributions of this paper can be summarized as follows:

\begin{itemize}
\item We discover an uneven distribution of average gradient norms across layers during LoRA fine-tuning, suggesting that different layers hold varying levels of importance in mobile VLMs training.

\item We propose a hierarchical rank scheduling by allocating varying ranks to different layers in the mobile VLMs during LoRA fine-tuning. This method encompasses two optimizations: the coarse-grained approach categorizes layers based on average gradient norms and assigns an identical rank to layers within each group, while the fine-grained approach adjusts the ranks of components within individual layers.

\item We develop an end-to-end automated optimization process using a lightweight performance model, which dynamically determines the optimal rank for each layer during the instruction tuning of mobile VLMs.
\item HyDRA surpasses LoRA across the board in instruction tuning on mobile VLMs, achieving a 4.7\% improvement on MME ~\cite{fu2024mme} and a 4.1\% improvement on MMB ~\cite{liu2023mmbench}. HyDRA even outperforms full-parameter fine-tuning under certain benchmarks.
\end{itemize}

\section{Background and Motivation}
\subsection{Low-Rank Adaptation} 
LoRA efficiently fine-tunes large pre-trained models by updating only a subset of their parameters. It introduces low-rank matrices in each layer to capture task-specific information, decomposed into matrices $A$ and $B$, where \(A \in \mathbb{R}^{d \times r} \) and \(B\in\mathbb{R}^{r \times d} \). Here, $r$ represents the rank, a hyperparameter that remains consistent across all layers. This reduces the number of updated parameters, lowering  computational costs. LoRA maintains the pre-trained model's structure while integrating new capabilities, making it ideal for fine-tuning large models in resource-constrained situations.

\subsection{Instruction Tuning for VLMs}
The entire training process of the VLMs consists of two stages: pre-training and instruction tuning~\cite{zhang2024mm}.  During the first step, the vision encoder and LLMs are typically frozen, with the training focused solely on the efficient projector. Subsequently, instruction tuning is applied to both the projector and LLMs to enhance visual understanding and expression capabilities by refining the model using a language modeling loss function. During the instruction tuning phase, we can choose either full-parameter fine-tuning or LoRA fine-tuning. Full-parameter fine-tuning yields better performance but requires substantial computational resources and is prone to overfitting, while LoRA fine-tuning reduces graphics memory consumption, providing an efficient method for training mobile VLMs.

%

\subsection{Motivation}
To explore how LoRA can be effectively utilized to train VLMs and enhance its performance, we conduct empirical studies on MobileVLM, including an analysis of the average gradient norms across different layers. In deep learning models, the average gradient norm can be mathematically defined as the mean norm of all parameter gradients.
Suppose we have a model with a parameter set \(\theta = \{\theta_1, \theta_2, \ldots, \theta_n\}\), where each parameter \(\theta_i\) has a corresponding gradient \(\nabla \theta_i\). The formula for the average gradient norms of layer $\ell$ is as follows:
\begin{equation}
w^{(l)} = \frac{1}{n} \sum_{i=1}^{n} \|\nabla \theta_i^{(l)}\|.
\end{equation}

We carefully track the average gradient norms of each layer following updates, as shown in Fig. 3. This visualization highlights a prominent trend: in LoRA fine-tuning, as the layers go deeper in the model, the average gradient norms generally increase.

\begin{figure}[t]
\centering
\includegraphics[width=0.8\linewidth]{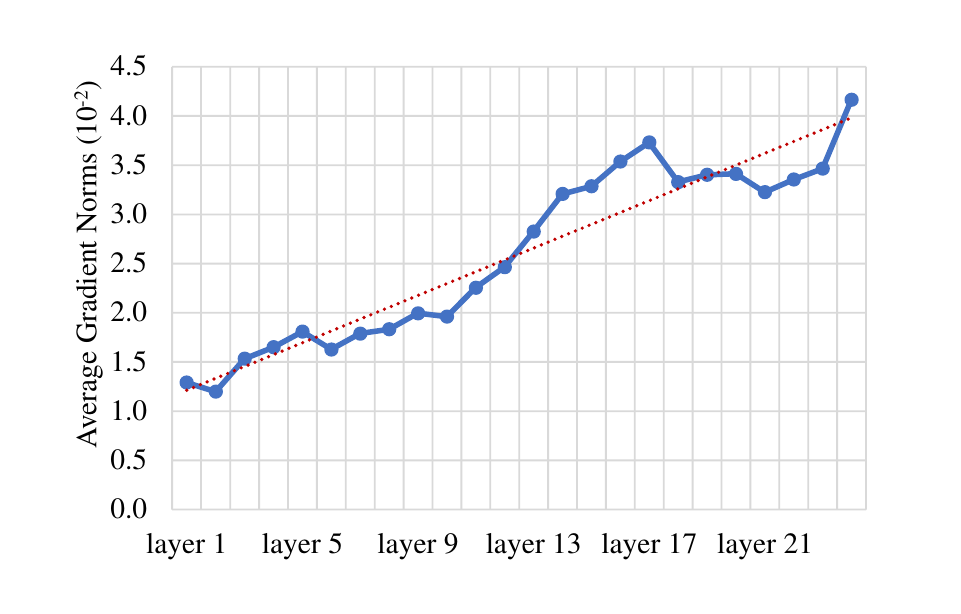} 
\caption{Average gradient norms of each layer with LoRA. }
\end{figure}

When we analyze the components within each layer of the model, we focus on several key projections. These include the query (${Q}$), key (${K}$), value (${V}$), and output (${O}$) projections of the attention matrix. We also examine the up (${Up}$), down (${Down}$), and gate ($Gate$) projections of the FFN. We observe distinct differences in their average gradient norms. Specifically, the average gradient norms for $Q$ and ${K}$ are relatively small, whereas those for ${Up}$ are relatively large, as shown in Table \uppercase\expandafter{\romannumeral1}.

The fluctuations in average gradient norms indicate the degree of variation in parameter updates during training. Thus, we hypothesize that the rank of layers within LoRA can dynamically adapt in response to increases in average gradient norms during the training of mobile VLMs. Additionally, the rank of the internal components within each layer should also be adjusted dynamically. This approach is anticipated to enhance the model's performance.

\begin{table}[t]
\centering
\caption{{Average gradient norms of components with LoRA}}
\begin{tabular}{l c c c c c c c}
\toprule
& \multicolumn{7}{c}{\textbf{Components of Layers}} \\ \cmidrule(lr){2-8}
 & \textbf{Down} & \textbf{Gate }& \textbf{Up} & \textbf{K }& \textbf{O} & \textbf{Q} &\textbf{ V }\\ \midrule
Value & 1.49 & 1.47 & \textbf{1.67} & \textbf{0.64} & 1.43 & \textbf{0.63} & 1.42 \\
\bottomrule
\end{tabular}
\label{tab:gradient_norms}
\end{table}

\subsection{An Example of Exploration} Based on the above hypothesis, we make corresponding adjustments to the rank value in LoRA fine-tuning. We find that models trained with increasing rank values indeed outperform those trained with fixed rank values. The specific performance results are shown in Table \uppercase\expandafter{\romannumeral2}.

\begin{table}[t]
\centering
\caption{{An example with increasing rank. The rank of \textit{Case} is structured as follows: the first 8 layers each have a rank of 64, the middle 8 layers each have a rank of 128, and the last 8 layers each have a rank of 256.}}
\begin{tabular}{l c c c c c c}
\toprule
\textbf{Rank }& \textbf{SQA\textsuperscript{I} }& \textbf{GQA} &\textbf{ VQA\textsuperscript{T} }& \textbf{POPE} & \textbf{MMB} &\textbf{ MME} \\ \midrule
128        & 51.95 & 55.28 & 40.28 & 83.76 & 46.56 & 1147.0 \\
\textit{Case} & \textbf{52.42} & \textbf{55.65} & \textbf{40.70} & 83.59 & \textbf{47.16} & \textbf{1191.9} \\
256        & \textbf{52.42} & 55.50 & 40.29 & \textbf{84.0} & 45.70 & 1130.1 \\ 
\bottomrule
\end{tabular}
\label{tab:increasing_rank}
\end{table}

\section{Method}
\subsection{Problem Definition}
In this section, we introduce some important terminology used in our method.

\textbf{Computation Graph.} We can define a LLM by a computation graph \( G \). The \( G \) is a directed acyclic graph. In this paper, we denote the first layer of the model as Layer 1, continuing sequentially up to the final layer.  Each layer in the \( G \) can be divided into several components such as \( Q \), \( K \), \( V \) of the attention matrix. 


\textbf{Stage.}
Based on the variation in the average gradient norms, we partition the computation graph \( G \) into \( t \) stages: \( S_1, S_2, \ldots, S_t \). Each stage contains a few layers, which have similar average gradient norms, as shown in Fig. 4.

\begin{figure}[t]
\centering
\includegraphics[width=0.8\linewidth]{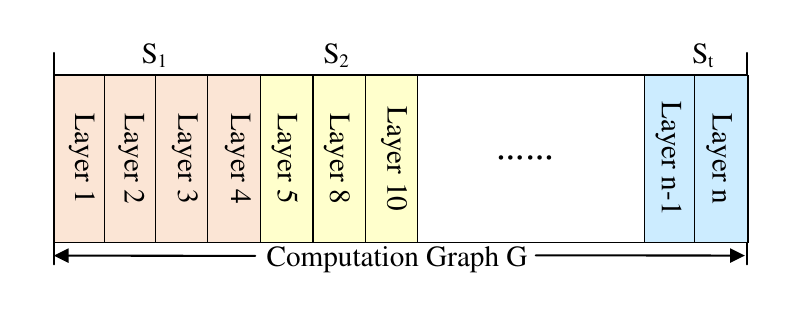} 
\vspace{-1em} 
\caption{Stage partitioning in the computation graph. }
\end{figure}

\textbf{Schedule.}
After obtaining the aforementioned $G$ and $Stage$, the challenge is how to divide the stages to achieve better model performance. To address this issue, We define a schedule \( Z \) for the \( G \) as :
\begin{equation}
Z = \{(S_1, R_1), (S_2, R_2), \ldots  (S_i, R_i),\ldots  (S_t, R_t)\} ,
\end{equation}
where \( S_i \) is the \( i \)-th stage which includes the set of LLM  layers, and \( R_i \) is the corresponding rank for \( S_i \) stage. 

Assume that in the \( i \)-th layer, the attention mechanism's components \( Q, K, V,\) and \(O \) each have their own ranks, and similarly, the $Up, Down$ and $Gate$ projections of FFN also have their respective ranks. We represent \( R_i \) as the rank of \( i \)-th layer as follows:
\begin{equation}
R_i = \{R_{i}^Q, R_{i}^K, R_{i}^V, R_{i}^O, R_{i}^{Up}, R_{i}^{Down}, R_{i}^{Gate}\} .  
\end{equation}
If all rank values within \( R_i \) are identical, it is referred to as \textbf{coarse-grained rank}. If the rank values within \( R_i \) are different, such as \( R^   Q_{i} \neq R^V_{i} \), it is referred to as \textbf{fine-grained rank}.

\begin{figure*}[t]
\centering
\includegraphics[width=0.8\linewidth]{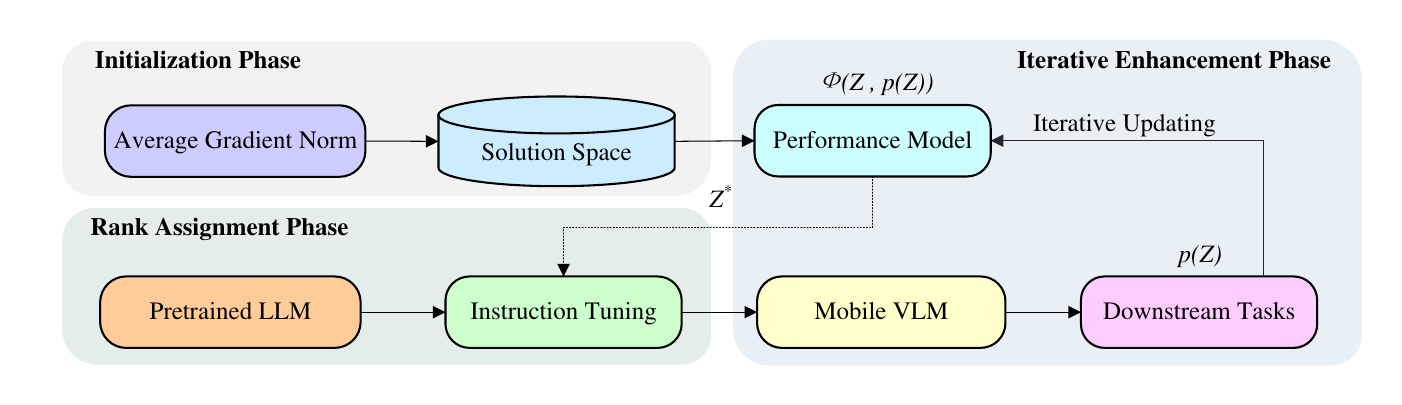} 
\caption{An end-to-end learning-based framework for optimizing hierarchical and dynamic ranks in
mobile VLMs instruction tuning. \textbf{Initialization Phase}: Define the solution space using average gradient norms. \textbf{Rank Assignment Phase}: Assign ranks and conducts instruction tuning accordingly. \textbf{Iterative Enhancement Phase}: Evaluate mobile VLMs on downstream tasks to obtain $p(Z)$, then iteratively refine the performance model $\Phi$ and predict the optimal rank scheduling $Z^*$.}
\end{figure*}

\textbf{Formulation.} Based on the definitions of the important terminology mentioned above, we incorporate these terms into the LoRA fine-tuning. The function \( g(G; Z) \) is defined to quantify the number of training parameters of  \( G \) following schedule \( Z \). 
Let \( p(Z) \) be the set of performance results for the trained model.
 We aim to find a schedule \( Z^* \) to maximize \( p \) and ensure that function \( g(G; Z) \) remains below a specified value. This can be directly formulated as a constrained optimization problem as follows:
\begin{equation}
 Z^* = \arg\max_{Z} \, p(Z) \quad \text{subject to} \quad g(G, Z) \leq C .
 \end{equation}

Here, $C$ represents a specified constant threshold, defined as the number of parameters required for training mobile VLMs using standard LoRA fine-tuning.


\subsection{HyDRA Framework}
We propose a learning-based framework, HyDRA, which consists of three critical phases, as depicted in Fig. 5.

\textbf{Initialization Phase.} 
According to Fig. 3, we define the solution space based on the trend of changes in average gradient norms. \(\Delta d\) is a positive integer variable. $N_s$ represents the number of stages. We define \( R_{i} \) as:
\begin{equation}
R_{i+1} \geq R_{i} + \Delta d, \quad \Delta d, i=1,2, \ldots, N_s,
\end{equation}
where \( R_{i} \) is a positive  integer, $R_{i}^{(l)}$ represents the rank value of layer $l$ in stage $i$. Let $N_{l}$ denote the total number of layers in the model. We define \(N_{e}\) as the number of layers in each stage. $R_{standard}$ represents the rank value of the standard LoRA. The total number of ranks is denoted as $N_R$ :
\begin{equation}
N_R=\sum_{i=1}^{N_s} \sum_{l=1}^{N_{e}} R_{i}^{(l)} \leq   R_{\text{standard}} \times N_l,  R_{i}^{(l)} \in \mathbb{Z}^+.
\end{equation}

\textbf{Rank  Assignment Phase.}
Within the constraints of the solution space, we define an adjustment factor $\alpha$, determined by the number of model layers, it is calculated as $\alpha= \frac{N_{l} - 1}{2}$. Therefore, the rank value for the first stage of initialization  $R_1^{initial}$ is denoted as:
\begin{equation}
R_{1}^{initial} = 2 \left\lfloor \frac{ R_{\text{standard}} - 2\alpha}{2} \right\rfloor .
\end{equation}

We can divide the stages based on the average gradient norms. Firstly, we need to specify the number of stages $N_s$ to partition $G$, ensuring that the rank value within each stage remains consistent. After specifying 
$N_s$, we apply the $K$-means clustering algorithm~\cite{hartigan1979algorithm} to group the layers based on their average gradient norms, determining which layers belong to which cluster that represents the stage. Subsequently, we sort the stages based on their values.

To ensure a more uniform smoothing of ranks across each stage, we distribute the rank values to different stages in a linearly increasing manner, starting from $R_1^{initial}$. If \( N_{e}^{(S_t)} \) represents the number of layers in the last stage \( S_t \), then the rank values corresponding to each stage are given by:
\begin{equation}
R_i = 
\begin{cases} 
\left\lfloor \frac{N_R - \sum_{j=1}^{N_s-1} \sum_{l=1}^{N_{e}} R_{j}^{(l)}}{N_{e}^{(S_t)}} \right\rfloor & \text{if } i = N_s \\
R_1 +(i-1)\cdot \Delta d & \text{otherwise}
\end{cases} .
\end{equation}

We use $i$ to represent the number of iterations, $N_{e}^{(S_i)}$ represtens the number of layers in the $i$-th stage. We set $N_{remain}$ to represent the remaining number of ranks, where 
 $N_{remain}=N_R-\sum_{j=1}^{i}({R_j}\cdot N_{e}^{(S_j)})$. Iteratively apply the rank allocation method described above. After each iteration, increment \( R_{\text{1}} \) by 1. Continue the iterations until \( N_{remain}\)  becomes less than 0 or \( R_t \leq R_{t-1} \). At this point, the iteration stops, and the allocation of ranks from the previous iteration is retained for all layers. Thus, we obtain the coarse-grained rank of the model. And we use $|{R_i}|$ to represent the coarse-grained rank value of the $i$-th stage.

According to Table \uppercase\expandafter{\romannumeral1}, we believe that the rank values of $R_i^K$ and $R_i^Q$  should be reduced, while $R_i^{Up}$ should be increased. Therefore, we appropriately adjust the rank values of components within each layer. We set $R_i^K$, $R_i^Q$, and $R_i^{Up}$ as follows:
\begin{gather}
R_{i}^Q = R_{i}^K = |R_{i}| - \Delta d , \\
R_{i}^{Up} = |R_{i}| + \Delta d ,
\end{gather}
resulting in a set $Z$. Thus, the fine-grained rank of the model is obtained.

\begin{table*}[ht] 
\centering
\caption{A comparison of various methods and their corresponding performance metrics. The \textit{Res.} column represents the image resolution of the vision model. The \textit{PT} and \textit{IT} columns indicate the data sizes during the pretraining stage and the instruction tuning stage, respectively. The term \textit{w/ LoRA} denotes instruction tuning using LoRA, while \textit{w/ Full} signifies instruction tuning through full-parameter fine-tuning.}
\setlength{\tabcolsep}{3.5pt} 
\renewcommand{\arraystretch}{1.2} 
\resizebox{\textwidth}{!}{%
\begin{tabular}{@{}llccc|cccccc@{}}
\toprule
\textbf{Method} & \textbf{LLM} & \textbf{Res.} & \textbf{PT} & \textbf{IT} & \textbf{MME} & \textbf{MMB} & \textbf{VQA}\textsuperscript{\textbf{T}} & \textbf{POPE} & \textbf{GQA} & \textbf{SQA}\textsuperscript{\textbf{I}} \\ 
\midrule
BLIP-2 & Vicuna-13B & 224 & 129M & - & 1293.8 & - & 42.5 & 85.3 & 41.0 & 61.0 \\
MiniGPT-4 & Vicuna-7B & 224 & 5M & 5K & 581.7 & 23.0 & - & - & 32.2 & - \\
InstructBLIP & Vicuna-7B & 224 & 129M & 1.2M & - & 36.0 & 50.1 & - & 49.2 & 60.5 \\
InstructBLIP & Vicuna-13B & 224 & 129M & 1.2M & 1212.8 & - & 50.7 & 78.9 & 49.5 & 63.1 \\
IDEFICS-9B & LLaMA-7B & 224 & 353M & 1M & - & 48.2 & 25.9 & - & 38.4 & - \\
IDEFICS-80B & LLaMA-65B & 224 & 353M & 1M & - & 54.5 & 30.9 & - & 45.2 & - \\
Qwen-VL & Qwen-7B & 448 & 1.4B & 50M & 1487.6 & 38.2 & 63.8 & - & 59.3 & 67.1 \\
LLaVA-1.5 & Vicuna-7B & 336 & 558K & 665K & 1510.7 & 64.3 & 58.2 & 85.9 & 62.0 & 66.8 \\ 
\midrule
MobileVLM 1.7B w/ Full & MobileLLaMA 1.4B & 336 & 558K & 665K & 1176.85 & 53.35 & 40.36 & 83.29 & 56.51 & 55.08 \\
MobileVLM 1.7B w/ LoRA & MobileLLaMA 1.4B & 336 & 558K & 665K & 1147.00 & 46.56 & 40.28 & 83.76 & 55.28 & 51.95 \\
HyDRA(\textit{Fine-grained}) & MobileLLaMA 1.4B & 336 & 558K & 665K & 1200.88 & 48.45 & 40.75 & 83.97 & 55.64 & 51.78 \\
HyDRA(\textit{Coarse-grained}) & MobileLLaMA 1.4B & 336 & 558K & 665K & 1200.91 & 47.77 & 40.98 & 84.51 & 55.33 & 52.2 \\ 
\midrule
MobileVLM 3B w/ Full & MobileLLaMA 2.7B & 336 & 558K & 665K & 1296.54 & 59.71 & 48.58 & 84.51 & 59.03 & 57.93 \\
MobileVLM 3B w/ LoRA & MobileLLaMA 2.7B & 336 & 558K & 665K & 1256.38 & 57.22 & 46.58 &  84.14 & 58.13 & 56.85 \\
HyDRA(\textit{Fine-grained}) & MobileLLaMA 2.7B & 336 & 558K & 665K &1261.41  &57.99  & 46.91 & 84.66 &  58.58& 57.39 \\ 
HyDRA(\textit{Coarse-grained})  & MobileLLaMA 2.7B & 336 & 558K & 665K & 1298.83  &  58.93 & 47.14 & 83.77 &  58.37& 56.94 \\ 
\bottomrule
\end{tabular}%
}
\label{tab:comparison_metrics}
\end{table*}

\textbf{Iterative Enhancement Phase.} We construct a performance model to estimate the effectiveness of mobile VLMs on various downstream tasks. This model uses an iterative process to continuously update its weights.
 Within a constrained solution space, we explore various $Z$, train the respective mobile VLMs, apply these models to downstream tasks, and collect feedback. This set of $Z$ and their respective evaluation values serve as incremental data to refine the performance model, which in turn predicts a more suitable $Z$. 

When $g(G;Z) \leq C$ and the requirements for the maximum number of iterations are satisfied, we iteratively utilize the performance model to find the optimal $Z^*$. We then train the corresponding mobile VLM, which achieves the optimal $p(Z^*)$ on downstream tasks.

\section{Experiments}

\subsection{Experimental Setup}

\textbf{LLMs and Benchmarks.}
We train VLMs using MobileLLaMA 1.4B and MobileLLaMA 2.7B, which are downscaled versions of LLaMA~\cite{chu2023mobilevlm}. We evaluate the performance of mobile VLMs on GQA~\cite{hudson2019gqa}, \text{SQA}\textsuperscript{I}~\cite{lu2022learn}, \text{VQA}\textsuperscript{T}~\cite{singh2019towards}, POPE~\cite{li2023evaluating}, MME, and conduct comprehensive comparisons with MMB. 


\textbf{Implementation Details.}
We use the following software and hardware configurations: PyTorch version 2.3.0; Transformers library version 4.41.0; PEFT (Parameter-Efficient Fine-Tuning) library version 0.11.1; CUDA version 12.1; and 4 NVIDIA Tesla A100 GPUs with 80 GB of memory each.
Instruction tuning is performed on the LLM and the projector using the LLaVA-Instruct-158K dataset ~\cite{liu2024visual} for one epoch, with a learning rate of $1 \times 10^{-4}$ for MobileLLaMA 1.4B and  $2 \times 10^{-4}$ for MobileLLaMA 2.7B. We use a batch size of 32 per device, resulting in a global batch size of 128.
We choose the AdamW optimizer ~\cite{loshchilov2017decoupled} with no weight decay and a cosine learning rate schedule with a warmup ratio of 3\%. The random seed is set to 42 in experiments.

The proposed performance model employs one encoder layer of Transformer~\cite{vaswani2017attention} with 4 multi-head attention heads and the model dimension is 32. The training data comprises all information necessary to cover rank configuration. The label data consists of six benchmark evaluation metrics corresponding to the training data. We use the L2 loss function and utilize the Adam~\cite{kingma2014adam} optimizer for optimization.

\subsection{Main Results.}

\textbf{Baseline and Configurations.} We use the MobileVLM as the base model and employ three training methods: full-parameter fine-tuning, LoRA, and HyDRA to train the model. In experiments,  there are two baselines: one is standard LoRA with the rank set to 128, and the other is full-parameter fine-tuning. To ensure a fair comparison, we establish a consistent experimental environment.
\begin{table*}[t] 
    \centering
    \caption{Results of the ablation study on coarse-grained and fine-grained ranks across six tasks.}
    \setlength{\tabcolsep}{6pt} 
    \renewcommand{\arraystretch}{1.2} 
    \resizebox{\textwidth}{!}{%
    \begin{tabular}{lcccc|ccccc}
        \toprule
        & \multicolumn{4}{c|}{ \textbf{Coarse-grained Adjustment of Rank} }& \multicolumn{5}{c}{ \textbf{Fine-grained Adjustment of Rank} }\\
        \cmidrule(lr){2-5} \cmidrule(lr){6-10}
        \textbf{Benchmark} & \textbf{Config1} & \textbf{Config2} & \textbf{Config3} & \textbf{Config4} & \textbf{Setting1} & \textbf{Setting2} & \textbf{Setting3} & \textbf{Setting4} & \textbf{Setting5} \\ 
        \midrule
        MME & 1162.3 & 1183.71 & 1191.91 & 1200.91 & 1200.88 & 1176.66 & 1181.88 & 1171.83 & 1155.82 \\
        GQA & 55.49 & 55.36 & 55.65 & 55.33 & 55.64 & 55.67 & 55.45 & 55.42 & 55.56 \\
        \text{VQA}\textsuperscript{T} & 40.59 & 40.12 & 40.7 & 40.98 & 40.75 & 40.34 & 40.7 & 41.16 & 41.15 \\
        POPE & 83.67 & 83.66 & 83.59 & 84.51 & 83.97 & 84.08 & 83.46 & 84.41 & 83.96 \\
        MMB & 45.53 & 45.53 & 47.16 & 47.77 & 48.45 & 48.02 & 48.11 & 48.88 & 48.28 \\
        \text{SQA}\textsuperscript{I} & 52.7 & 52.44 & 52.42 & 52.2 & 51.78 & 52.2 & 52.75 & 52.65 & 52.68 \\
        \bottomrule
    \end{tabular}%
    }
    \label{tab:ablation_study}
\end{table*}

\textbf{Experimental Results.}
We present the performance of the trained VLMs using full-parameter fine-tuning, LoRA, and HyDRA, with the best results for each task presented in Table \uppercase\expandafter{\romannumeral3}. The experiments are conducted on models of different parameter sizes.

 On MobileLLaMA 1.4B, we compare HyDRA with both full-parameter fine-tuning and standard LoRA. HyDRA outperforms the full-parameter fine-tuning on tasks such as MME, \text{VQA}\textsuperscript{T}, and POPE. For instance, HyDRA (Fine-grained) achieves a metric of 1200.88 compared to 1176.85 on MMB. HyDRA consistently surpasses LoRA across all six benchmarks. For example, HyDRA (Fine-grained) achieves a metric of 1200.88 compared to 1147.00 on MME, and outperforms LoRA by 4.1\% on MMB. HyDRA (Coarse-grained) outperforms LoRA by 4.7\% on MME. 
Meanwhile, on MobileLLaMA 2.7B, HyDRA outperforms full-parameter fine-tuning on tasks such as MME and POPE. HyDRA consistently surpasses LoRA across all six benchmarks. Notably, HyDRA (Coarse-grained) outperforms LoRA by 3.4\% on MME and by 3\% on the MMB.

\begin{figure}[t]
\centering
\includegraphics[width=0.8\linewidth]{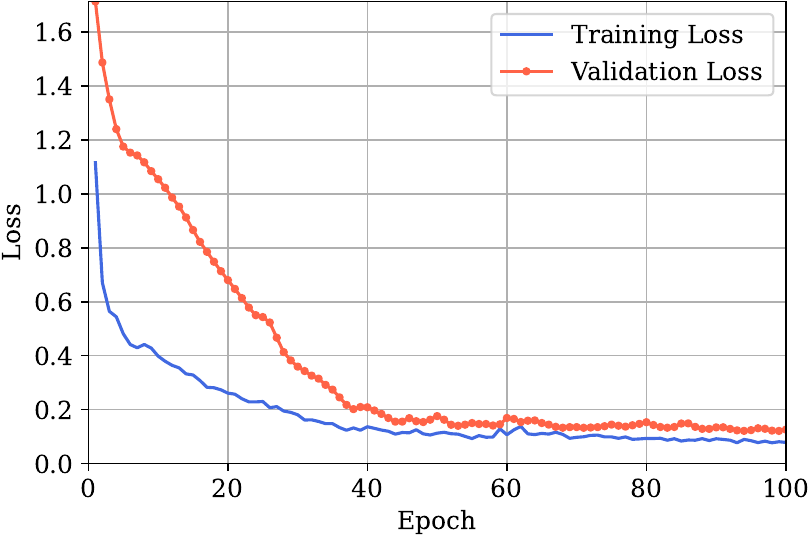} 
\caption{Comparison of training and validation loss of the performance model across epochs. }
\end{figure}
All results from the main experiment are obtained using the performance model in HyDRA. The validation results from the trained mobile VLM are obtained and utilized to develop an expanded dataset. The dataset is divided into training and validation sets with a ratio of 8:2. The convergence of the performance model is depicted in Fig. 6. The curves of various colors represent the convergence of the model across training and validation datasets.
Overall, the experimental results demonstrate that HyDRA consistently outperforms LoRA across all tasks, and even surpasses full-parameter fine-tuning on MME, \text{VQA}\textsuperscript{T}, and POPE.


\subsection{Ablation Study}
We utilize MobileLLaMA 1.4B, the CLIP-ViT-L/14@336px model~\cite{radford2021learning} and HyDRA in experiments.

\textbf{Coarse-grained  Rank.} 
The experiments are conducted with various rank values across different layers, the results are shown in Table \uppercase\expandafter{\romannumeral4}. We set the rank values for the ablation experiment as shown in Table \uppercase\expandafter{\romannumeral5}. 

\begin{table}[t]
    \centering
    \caption{Configurations for coarse-grained rank adjustment.}
    \label{tab:rank_distribution}
    \renewcommand{\arraystretch}{1.2} 
    \setlength{\tabcolsep}{5pt} 
    \begin{tabular}{lcc}
        \toprule 
        \textbf{Rank Setup} & \textbf{Layers} & \textbf{Rank Value} \\ 
        \midrule 
        Config1 & [1–4, 5–20, 21–24] & (128, 64, 128) \\
        Config2 & [1–4, 5–20, 21–24] & (256, 128, 256) \\
        Config3 & [1–8, 9–16, 17–24] & (64, 128, 256) \\
        Config4 & [1–8, 9–12, 13–24] & (124, 126, 131) \\
        \bottomrule 
    \end{tabular}
\end{table}

For example, Config1 is divided into three stages: the first stage comprises layers 1-4 with a rank value of 128, the second stage comprises layers 5-20 with a rank value of 64, and the third stage comprises layers 21-24 with a rank value of 128. Config4 leverages HyDRA with a stage set to 3.This setting is designed for a fair comparison.

Config1 and Config2 share the same stage, but the rank value of Config2 is twice that of Config1, resulting in decreased performance on GQA, \text{VQA}\textsuperscript{T}, POPE, and \text{SQA}\textsuperscript{I}. This shows that increasing rank alone does not improve performance. Config3, with a progressively increasing rank distribution, outperforms Config2 on MME, GQA, \text{VQA}\textsuperscript{T}, and MMB, highlighting the benefit of a progressive rank distribution. Compared to Config3,  Config4 which uses HyDRA, shows further improvements on MME, \text{VQA}\textsuperscript{T}, POPE, and MMB, indicating that HyDRA significantly enhances model performance.

\textbf{Fine-grained Rank.}
The experiments are conducted with various rank values across components within the layers, the results are shown in Table \uppercase\expandafter{\romannumeral4}. According to HyDRA, we consistently set the stage to 3, resulting in $R_1 = 124$, $R_2 = 126$, and $R_3 = 131$. Additionally, we set $\Delta d$ equals 2. The settings of fine-grained ranks are shown as Table \uppercase\expandafter{\romannumeral6}.

\begin{table}[t]
    \centering
    \caption{Settings for fine-grained rank adjustment.}
    \renewcommand{\arraystretch}{1.5} 
    \setlength{\tabcolsep}{15pt} 
    \begin{tabularx}{0.6\columnwidth}{@{}lX@{}}
        \toprule
        \textbf{Setting} & \multicolumn{1}{c}{\textbf{Adjustment}} \\
        \midrule
        Setting1 & \begin{tabular}[t]{@{}l@{}}$R^Q_i, R^K_i \rightarrow |R_i| - \Delta d,$ \\ $R^{Up}_i \rightarrow |R_i| + \Delta d$\end{tabular} \\
        Setting2 & \begin{tabular}[t]{@{}l@{}}$R^Q_i, R^K_i \rightarrow |R_i| - \Delta d,$ \\ $R^{Up}_i, R^{Down}_i \rightarrow |R_i| + \Delta d$\end{tabular} \\
        Setting3 & $R^Q_i, R^K_i \rightarrow |R_i| - \Delta d$ \\
        Setting4 & \begin{tabular}[t]{@{}l@{}}$R^Q_i, R^K_i \rightarrow |R_i| - \frac{\Delta d}{2},$ \\ $R^{Up}_i \rightarrow |R_i| + \Delta d$\end{tabular} \\
        Setting5 & \begin{tabular}[t]{@{}l@{}}$R^Q_i, R^K_i \rightarrow |R_i| - \Delta d,$ \\ $R^{Up}_i \rightarrow |R_i| + 2\Delta d$\end{tabular} \\
        \bottomrule
    \end{tabularx}
    \label{tab:configurations}
\end{table}


Comparing Setting1 with Setting5 reveals that excessively increasing the \( R_i^{up} \) value degrades performance on MME, GQA, POPE, and MMB tasks, notably reducing MME by 3.9\%. In contrast, Setting3 shows that not increasing \( R_i^{up} \) causes performance drops across five tasks, excluding \text{SQA}\textsuperscript{I}. Setting4 indicates performance declines in MME, GQA, and \text{VQA}\textsuperscript{T}. Lastly, Setting2 demonstrates that increasing \( R^{down}_i \)  degrades performance on MME, \text{VQA}\textsuperscript{T}, and MMB but improves GQA, POPE, and \text{SQA}\textsuperscript{I} tasks. Overall, Setting1 produces the best overall results.

\textbf{Stage.} 
We conduct ablation experiments to assess the impact of stage sizes by comparing performance for $ N \in \{7, 11, 13, 17\}$. The results are summarized in Table \uppercase\expandafter{\romannumeral7}. Our findings indicate that decreasing the value of $\Delta d$, when $N=17$, results in poorer performance on the benchmarks. However, variations in stage sizes do not lead to a significant pattern in model performance.

\begin{table}[t]
\centering
\caption{Results of the ablation study on stage size.}
\setlength{\tabcolsep}{3pt}  
\renewcommand{\arraystretch}{1.2} 
\begin{tabular}{lccccc} 
\toprule 
\multicolumn{1}{c}{} & \multicolumn{5}{c}{ \textbf{Stage Size}} \\ 
\cmidrule(lr){2-6}
\multicolumn{1}{c}{ \textbf{Benchmark}} &
\multicolumn{1}{c}{ \textbf{N=7}} & 
\multicolumn{1}{c}{ \textbf{N=11}} & 
\multicolumn{1}{c}{ \textbf{N=13}} & 
\multicolumn{1}{c}{ \textbf{N=17}} & 
\multicolumn{1}{c}{ \textbf{N=17}} \\ 
\multicolumn{1}{c}{} &
\multicolumn{1}{c}{\boldmath{$\Delta d=2$}} & 
\multicolumn{1}{c}{\boldmath{$\Delta d=2$}} & 
\multicolumn{1}{c}{\boldmath{$\Delta d=2$}} & 
\multicolumn{1}{c}{\boldmath{$\Delta d=2$} }& 
\multicolumn{1}{c}{\boldmath{$\Delta d=1$}} \\ 
\midrule
MME   & 1176.46 & 1172.30 & 1187.20 & 1183.65 & 1164.59 \\
GQA   &  55.68  &  55.42  &  55.80  &  55.14  &  55.45  \\
\text{VQA}\textsuperscript{T}&  40.60  &  40.48  &  40.64  &  40.48  &  40.46  \\
POPE  &  84.03  &  83.66  &  83.91  &  84.28  &  83.84  \\
MMB   &  47.08  &  47.59  &  44.85  &  46.74  &  45.62  \\
\text{SQA}\textsuperscript{I} &  53.15  &  52.61  &  52.91  &  52.82  &  52.68  \\
\bottomrule 
\end{tabular}
\end{table}

\textbf{Performance Model.}
In experiments, we evaluate four architecture designs for performance model: multi-layer perceptron (MLP), LightGBM, random forest (RF), and Transformer encoder (Ours), using two key metrics: mean squared error (MSE) and mean absolute error (MAE). The results, summarized in Table \uppercase\expandafter{\romannumeral8}. Transformer consistently outperforms the others with the lowest MSE of 0.1257 and the lowest MAE of 0.2531, indicating superior predictive accuracy and robustness. In contrast, MLP and RF perform similarly, with MSE values of 0.3813 and 0.3526, and MAE scores of 0.4617 and 0.4619. LightGBM also performs well, with an MSE of 0.3612 and an MAE of 0.3228. These findings suggest that our method provides the most reliable performance in terms of both error metrics.

\begin{table}[t]
\centering
\caption{Results of the ablation study on performance model.}
\setlength{\tabcolsep}{3pt} 
\renewcommand{\arraystretch}{1.2} 
\begin{tabular}{lcccc} 
\toprule
\multicolumn{1}{c}{} & \multicolumn{4}{c}{\textbf{Performance Model} }\\ 
\cmidrule(lr){2-5}
\textbf{Metrics} & \textbf{MLP} & \textbf{LightGBM} & \textbf{RF} & \textbf{Ours} \\
\midrule
MSE   & 0.3813 & 0.3612 & 0.3526 & 0.1257 \\
MAE   & 0.4617 & 0.3228 & 0.4619 & 0.2531 \\
\bottomrule
\end{tabular}
\end{table}

\section{Related Work}

\textbf{Large Language Model.} With the launch of ChatGPT, the technology of LLMs has rapidly advanced, attracting widespread attention and application. The popular open-source LLMs such as GLM~\cite{du2022glm}, LLaMA~\cite{touvron2023llama} and Mistral~\cite{jiang2023mistral} has greatly promoted iterative innovation in LLMs technology. However, due to the enormous training costs of LLMs, there is an increasingly trend towards researching small-scale alternative solutions.
For example, models like TinyLLaMA~\cite{zhang2024tinyllama}, MiniCPM~\cite{hu2024minicpm}, and Phi~\cite{li2023textbooks} have parameter counts of less than 3 billion. These studies indicate that high-quality training data and feasible pre-training methods can enable small models to achieve comparable capabilities as LLMs.

\textbf{Vision Language Model.} In addition to text modality, the ability to process other modalities is also crucial for the model.
VLMs enhance the capability of LLMs in processing visual information, broadening the scope of application scenarios. OpenAI has developed GPT-4V~\cite{openai2023gpt}, demonstrating remarkable capabilities. To date, the community has contributed many exciting studies on VLMs.
Flamingo~\cite{alayrac2022flamingo}, bridges pretrained vision and language models, handling interleaved sequences of text and image data. MiniGPT4~\cite{zhu2023minigpt} aligns a frozen visual encoder with a frozen LLM, Vicuna~\cite{chiang2023vicuna}, using one projection layer. LLaVA~\cite{liu2024visual} is a large multimodal model, which connect a vision encoder and a LLM for comprehensive visual and language understanding.
However, the enormous parameter size of VLMs results in excessively high training costs. To address this issue, LLaVA-Phi~\cite{zhu2024llava} builds a multimodal model based on a small-scale LLM named Phi-2~\cite{javaheripi2023phi}.
TinyLLaVA~\cite{zhang2024tinyllama} introduces a training framework capable of integrating visual information with small-scale LLM, such as TinyLLaMA.
MobileVLM introduces MobileLLaMA, reducing the size of the LLaMA model through traditional neural compression techniques. Additionally, MobileVLM proposes a lightweight projector to further reduce the computational costs incurred by the visual encoding process, providing new opportunities for running VLMs on mobile devices.

\textbf{Parameter-Efficient Fine-Tuning.}
PEFT is one of the effective methods for addressing the enormous computational requirements of training VLMs.
Serial Adapter~\cite{houlsby2019parameter} enhances each Transformer block by adding two adapter modules.
Adapter modules produce a compact and extensible model by adding only a few trainable parameters per task.  
Prefix-tuning~\cite{li2021prefix} adds learnable vectors to Transformer keys and values, with an MLP generating these vectors during training. Only the prefix vectors are used at inference. AUTOPEFT~\cite{zhou2024autopeft} defines a search space with serial adapters, parallel adapters, and prefix tuning, and uses high-dimensional Bayesian optimization for efficient neural architecture search. It enhances PEFT configurations across various tasks. Overall, applying PEFT to train VLMs is a highly efficient strategy, leading to substantial gains in performance and reduced computational costs.

\section{Conclusion}
In this paper, we introduce HyDRA, a novel framework that enables instruction tuning for mobile VLMs. HyDRA addresses the limitations of standard LoRA by dynamically assigning ranks and adaptively adjusting ranks for each layer based on their average gradient norms. Additionally, HyDRA employs an end-to-end automatic optimization process with a lightweight performance model to determine and adjust ranks dynamically during instruction tuning. HyDRA consistently surpasses the baseline framework, achieving a 4.7\% improvement without increasing the number of training parameters. Looking ahead, we plan to explore the generalization of our method to other tasks, particularly within the video-language model domain.


\bibliographystyle{IEEEtran}
\bibliography{mybibliography}

@article{chiang2023vicuna,
  title={Vicuna: An open-source chatbot impressing gpt-4 with 90\%* chatgpt quality},
  author={Chiang, Wei-Lin and Li, Zhuohan and Lin, Zi and Sheng, Ying and Wu, Zhanghao and Zhang, Hao and Zheng, Lianmin and Zhuang, Siyuan and Zhuang, Yonghao and Gonzalez, Joseph E and others},
  journal={See https://vicuna. lmsys. org (accessed 14 April 2023)},
  volume={2},
  number={3},
  pages={6},
  year={2023}
}

@article{javaheripi2023phi,
  title={Phi-2: The surprising power of small language models},
  author={Javaheripi, Mojan and Bubeck, S{\'e}bastien and Abdin, Marah and Aneja, Jyoti and Bubeck, Sebastien and Mendes, Caio C{\'e}sar Teodoro and Chen, Weizhu and Del Giorno, Allie and Eldan, Ronen and Gopi, Sivakanth and others},
  journal={Microsoft Research Blog},
  year={2023}
}

@article{openai2023gpt,
  title={GPT-4V (ision) System Card},
  author={OpenAI, R},
  journal={Citekey: gptvision},
  year={2023}
}

@article{liu2024visual,
  title={Visual instruction tuning},
  author={Liu, Haotian and Li, Chunyuan and Wu, Qingyang and Lee, Yong Jae},
  journal={Advances in neural information processing systems},
  volume={36},
  year={2024}
}

@article{jiang2023mistral,
  title={Mistral 7B},
  author={Jiang, Albert Q and Sablayrolles, Alexandre and Mensch, Arthur and Bamford, Chris and Chaplot, Devendra Singh and Casas, Diego de las and Bressand, Florian and Lengyel, Gianna and Lample, Guillaume and Saulnier, Lucile and others},
  journal={arXiv preprint arXiv:2310.06825},
  year={2023}
}

@article{hu2024minicpm,
  title={Minicpm: Unveiling the potential of small language models with scalable training strategies},
  author={Hu, Shengding and Tu, Yuge and Han, Xu and He, Chaoqun and Cui, Ganqu and Long, Xiang and Zheng, Zhi and Fang, Yewei and Huang, Yuxiang and Zhao, Weilin and others},
  journal={arXiv preprint arXiv:2404.06395},
  year={2024}
}

@article{zhu2023minigpt,
  title={Minigpt-4: Enhancing vision-language understanding with advanced large language models},
  author={Zhu, Deyao and Chen, Jun and Shen, Xiaoqian and Li, Xiang and Elhoseiny, Mohamed},
  journal={arXiv preprint arXiv:2304.10592},
  year={2023}
}

@article{chu2023mobilevlm,
  title={Mobilevlm: A fast, reproducible and strong vision language assistant for mobile devices},
  author={Chu, Xiangxiang and Qiao, Limeng and Lin, Xinyang and Xu, Shuang and Yang, Yang and Hu, Yiming and Wei, Fei and Zhang, Xinyu and Zhang, Bo and Wei, Xiaolin and others},
  journal={arXiv preprint arXiv:2312.16886},
  year={2023}
}

@article{alayrac2022flamingo,
  title={Flamingo: a visual language model for few-shot learning},
  author={Alayrac, Jean-Baptiste and Donahue, Jeff and Luc, Pauline and Miech, Antoine and Barr, Iain and Hasson, Yana and Lenc, Karel and Mensch, Arthur and Millican, Katherine and Reynolds, Malcolm and others},
  journal={Advances in neural information processing systems},
  volume={35},
  pages={23716--23736},
  year={2022}
}

@article{zhang2024tinyllama,
  title={Tinyllama: An open-source small language model},
  author={Zhang, Peiyuan and Zeng, Guangtao and Wang, Tianduo and Lu, Wei},
  journal={arXiv preprint arXiv:2401.02385},
  year={2024}
}

@article{li2023textbooks,
  title={Textbooks are all you need ii: phi-1.5 technical report},
  author={Li, Yuanzhi and Bubeck, S{\'e}bastien and Eldan, Ronen and Del Giorno, Allie and Gunasekar, Suriya and Lee, Yin Tat},
  journal={arXiv preprint arXiv:2309.05463},
  year={2023}
}

@article{touvron2023llama,
  title={Llama: Open and efficient foundation language models},
  author={Touvron, Hugo and Lavril, Thibaut and Izacard, Gautier and Martinet, Xavier and Lachaux, Marie-Anne and Lacroix, Timoth{\'e}e and Rozi{\`e}re, Baptiste and Goyal, Naman and Hambro, Eric and Azhar, Faisal and others},
  journal={arXiv preprint arXiv:2302.13971},
  year={2023}
}

@inproceedings{du2022glm,
  title={GLM: General Language Model Pretraining with Autoregressive Blank Infilling},
  author={Du, Zhengxiao and Qian, Yujie and Liu, Xiao and Ding, Ming and Qiu, Jiezhong and Yang, Zhilin and Tang, Jie},
  booktitle={Proceedings of the 60th Annual Meeting of the Association for Computational Linguistics (Volume 1: Long Papers)},
  pages={320--335},
  year={2022}
}

@inproceedings{houlsby2019parameter,
  title={Parameter-efficient transfer learning for NLP},
  author={Houlsby, Neil and Giurgiu, Andrei and Jastrzebski, Stanislaw and Morrone, Bruna and De Laroussilhe, Quentin and Gesmundo, Andrea and Attariyan, Mona and Gelly, Sylvain},
  booktitle={International conference on machine learning},
  pages={2790--2799},
  year={2019},
  organization={PMLR}
}

@inproceedings{li2021prefix,
  title={Prefix-Tuning: Optimizing Continuous Prompts for Generation},
  author={Li, Xiang Lisa and Liang, Percy},
  booktitle={Proceedings of the 59th Annual Meeting of the Association for Computational Linguistics and the 11th International Joint Conference on Natural Language Processing (Volume 1: Long Papers)},
  pages={4582--4597},
  year={2021}
}

@article{hu2021lora,
  title={Lora: Low-rank adaptation of large language models},
  author={Hu, Edward J and Shen, Yelong and Wallis, Phillip and Allen-Zhu, Zeyuan and Li, Yuanzhi and Wang, Shean and Wang, Lu and Chen, Weizhu},
  journal={arXiv preprint arXiv:2106.09685},
  year={2021}
}

@article{zhou2024autopeft,
  title={Autopeft: Automatic configuration search for parameter-efficient fine-tuning},
  author={Zhou, Han and Wan, Xingchen and Vuli{\'c}, Ivan and Korhonen, Anna},
  journal={Transactions of the Association for Computational Linguistics},
  volume={12},
  pages={525--542},
  year={2024},
  publisher={MIT Press One Broadway, 12th Floor, Cambridge, Massachusetts 02142, USA~…}
}

@inproceedings{hudson2019gqa,
  title={Gqa: A new dataset for real-world visual reasoning and compositional question answering},
  author={Hudson, Drew A and Manning, Christopher D},
  booktitle={Proceedings of the IEEE/CVF conference on computer vision and pattern recognition},
  pages={6700--6709},
  year={2019}
}

@article{zhu2024llava,
  title={LLaVA-$\phi$: Efficient Multi-Modal Assistant with Small Language Model},
  author={Zhu, Yichen and Zhu, Minjie and Liu, Ning and Ou, Zhicai and Mou, Xiaofeng and Tang, Jian},
  journal={arXiv preprint arXiv:2401.02330},
  year={2024}
}

@inproceedings{singh2019towards,
  title={Towards vqa models that can read},
  author={Singh, Amanpreet and Natarajan, Vivek and Shah, Meet and Jiang, Yu and Chen, Xinlei and Batra, Dhruv and Parikh, Devi and Rohrbach, Marcus},
  booktitle={Proceedings of the IEEE/CVF conference on computer vision and pattern recognition},
  pages={8317--8326},
  year={2019}
}

@article{lu2022learn,
  title={Learn to explain: Multimodal reasoning via thought chains for science question answering},
  author={Lu, Pan and Mishra, Swaroop and Xia, Tanglin and Qiu, Liang and Chang, Kai-Wei and Zhu, Song-Chun and Tafjord, Oyvind and Clark, Peter and Kalyan, Ashwin},
  journal={Advances in Neural Information Processing Systems},
  volume={35},
  pages={2507--2521},
  year={2022}
}

@article{li2023evaluating,
  title={Evaluating object hallucination in large vision-language models},
  author={Li, Yifan and Du, Yifan and Zhou, Kun and Wang, Jinpeng and Zhao, Wayne Xin and Wen, Ji-Rong},
  journal={arXiv preprint arXiv:2305.10355},
  year={2023}
}

@article{fu2024mme,
  title={MME: A Comprehensive Evaluation Benchmark for Multimodal Large Language Models}, 
  author={Chaoyou Fu and Peixian Chen and Yunhang Shen and Yulei Qin and Mengdan Zhang and Xu Lin and Jinrui Yang and Xiawu Zheng and Ke Li and Xing Sun and Yunsheng Wu and Rongrong Ji},
  journal={arXiv preprint arXiv:2306.13394},
  year={2024}
}

@article{liu2023mmbench,
  title={Mmbench: Is your multi-modal model an all-around player?},
  author={Liu, Yuan and Duan, Haodong and Zhang, Yuanhan and Li, Bo and Zhang, Songyang and Zhao, Wangbo and Yuan, Yike and Wang, Jiaqi and He, Conghui and Liu, Ziwei and others},
  journal={arXiv preprint arXiv:2307.06281},
  year={2023}
}

@article{hartigan1979algorithm,
  title={Algorithm AS 136: A k-means clustering algorithm},
  author={Hartigan, John A and Wong, Manchek A},
  journal={Journal of the royal statistical society. series c (applied statistics)},
  volume={28},
  number={1},
  pages={100--108},
  year={1979},
  publisher={JSTOR}
}

@article{valipour2022dylora,
  title={Dylora: Parameter efficient tuning of pre-trained models using dynamic search-free low-rank adaptation},
  author={Valipour, Mojtaba and Rezagholizadeh, Mehdi and Kobyzev, Ivan and Ghodsi, Ali},
  journal={arXiv preprint arXiv:2210.07558},
  year={2022}
}

@article{dettmers2024qlora,
  title={Qlora: Efficient finetuning of quantized llms},
  author={Dettmers, Tim and Pagnoni, Artidoro and Holtzman, Ari and Zettlemoyer, Luke},
  journal={Advances in Neural Information Processing Systems},
  volume={36},
  year={2024}
}

@article{zhang2024vision,
  title={Vision-language models for vision tasks: A survey},
  author={Zhang, Jingyi and Huang, Jiaxing and Jin, Sheng and Lu, Shijian},
  journal={IEEE Transactions on Pattern Analysis and Machine Intelligence},
  year={2024},
  publisher={IEEE}
}

@article{zhang2024mm,
  title={Mm-llms: Recent advances in multimodal large language models},
  author={Zhang, Duzhen and Yu, Yahan and Li, Chenxing and Dong, Jiahua and Su, Dan and Chu, Chenhui and Yu, Dong},
  journal={arXiv preprint arXiv:2401.13601},
  year={2024}
}

@article{yin2023survey,
  title={A survey on multimodal large language models},
  author={Yin, Shukang and Fu, Chaoyou and Zhao, Sirui and Li, Ke and Sun, Xing and Xu, Tong and Chen, Enhong},
  journal={arXiv preprint arXiv:2306.13549},
  year={2023}
}

@article{huang2023lorahub,
  title={Lorahub: Efficient cross-task generalization via dynamic lora composition},
  author={Huang, Chengsong and Liu, Qian and Lin, Bill Yuchen and Pang, Tianyu and Du, Chao and Lin, Min},
  journal={arXiv preprint arXiv:2307.13269},
  year={2023}
}

@article{loshchilov2017decoupled,
  title={Decoupled weight decay regularization},
  author={Loshchilov, Ilya and Hutter, Frank},
  journal={arXiv preprint arXiv:1711.05101},
  year={2017}
}

@article{han2024parameter,
  title={Parameter-efficient fine-tuning for large models: A comprehensive survey},
  author={Han, Zeyu and Gao, Chao and Liu, Jinyang and Zhang, Sai Qian and others},
  journal={arXiv preprint arXiv:2403.14608},
  year={2024}
}

@inproceedings{radford2021learning,
  title={Learning transferable visual models from natural language supervision},
  author={Radford, Alec and Kim, Jong Wook and Hallacy, Chris and Ramesh, Aditya and Goh, Gabriel and Agarwal, Sandhini and Sastry, Girish and Askell, Amanda and Mishkin, Pamela and Clark, Jack and others},
  booktitle={International conference on machine learning},
  pages={8748--8763},
  year={2021},
  organization={PMLR}
}

@article{li2024vision,
  title={Vision-language models in remote sensing: Current progress and future trends},
  author={Li, Xiang and Wen, Congcong and Hu, Yuan and Yuan, Zhenghang and Zhu, Xiao Xiang},
  journal={IEEE Geoscience and Remote Sensing Magazine},
  year={2024},
  publisher={IEEE}
}

@article{luo2024cheap,
  title={Cheap and quick: Efficient vision-language instruction tuning for large language models},
  author={Luo, Gen and Zhou, Yiyi and Ren, Tianhe and Chen, Shengxin and Sun, Xiaoshuai and Ji, Rongrong},
  journal={Advances in Neural Information Processing Systems},
  volume={36},
  year={2024}
}

@article{kingma2014adam,
  title={Adam: a method for stochastic optimization},
  author={Kingma, DP},
  journal={arXiv preprint arXiv:1412.6980},
  year={2014}
}

@article{vaswani2017attention,
  title={Attention is all you need},
  author={Vaswani, Ashish},
  journal={arXiv preprint arXiv:1706.03762},
  year={2017}
}

@article{xu2023parameter,
  title={Parameter-efficient fine-tuning methods for pretrained language models: A critical review and assessment},
  author={Xu, Lingling and Xie, Haoran and Qin, Si-Zhao Joe and Tao, Xiaohui and Wang, Fu Lee},
  journal={arXiv preprint arXiv:2312.12148},
  year={2023}
}

\end{document}